\documentclass[10pt, a4paper]{article}
\usepackage{lrec2022} 

\usepackage{multibib}
\newcites{languageresource}{Language Resources}
\usepackage{graphicx}
\usepackage{tabularx}
\usepackage{soul}
\usepackage{microtype}
\usepackage{enumitem}
\usepackage{xspace}
\usepackage{xcolor}
\newcommand{\dataset}[0]{SuMe\xspace}
\newcommand{\eat}[1]{}

\usepackage{titlesec}
\titleformat{\section}{\normalfont\large\bfseries\center}{\thesection.}{1em}{}
\titleformat{\subsection}{\normalfont\SmallTitleFont\bfseries\raggedright}{\thesubsection.}{1em}{}
\titleformat{\subsubsection}{\normalfont\normalsize\bfseries\raggedright}{\thesubsubsection.}{1em}{}
\renewcommand\thesection{\arabic{section}}
\renewcommand\thesubsection{\thesection.\arabic{subsection}}
\renewcommand\thesubsubsection{\thesubsection.\arabic{subsubsection}}

\usepackage{epstopdf}
\usepackage[utf8]{inputenc}

\usepackage{hyperref}
\usepackage{xstring}

\usepackage{lipsum}
\usepackage{color}
\usepackage{authblk}

\newcommand{\edit}[1]{\textcolor{black}{#1}}

\title{SuMe: A Dataset Towards Summarizing Biomedical Mechanisms }

\name{Mohaddeseh Bastan\textsuperscript{$\spadesuit$}, Nishant Shankar\textsuperscript{$\clubsuit$}, Mihai Surdeanu\textsuperscript{$\heartsuit$}, \\ {\bf \large  Niranjan Balasubramanian\textsuperscript{$\spadesuit$}}}

\address{
\textsuperscript{$\spadesuit$}  Stony Brook University \hspace{15pt}
\textsuperscript{$\clubsuit$}Delft University of Technology \hspace{15pt}
\textsuperscript{$\heartsuit$} University of Arizona \\
         \{mbastan,  niranjan\}@cs.stonybrook.edu\\
         n.shankar@tudelft.nl \hspace{15pt}
          msurdeanu@email.arizona.edu\\
          }

\abstract{
Can language models read biomedical texts and explain the biomedical mechanisms discussed? In this work we introduce a biomedical mechanism summarization task.
Biomedical studies often investigate the mechanisms behind how one entity (e.g., a protein or a chemical) affects another in a biological context. The abstracts of these publications often include a focused set of sentences that present relevant supporting statements regarding such relationships, associated experimental evidence, and a concluding sentence that summarizes the mechanism underlying the relationship. We leverage this structure and create a summarization task, where the input is a collection of sentences and the main entities in an abstract, and the output includes the relationship and a sentence that summarizes the mechanism. Using a small amount of manually labeled mechanism sentences, we train a mechanism sentence classifier to filter a large biomedical abstract collection and create a summarization dataset with 22k instances. We also introduce conclusion sentence generation as a pretraining task with 611k instances. We benchmark the performance of large bio-domain language models. We find that while the pretraining task help improves performance, the best model produces acceptable mechanism outputs in only 32\% of the instances, which shows the task presents significant challenges in biomedical language understanding and summarization.
\medbreak
 }

\begin{document}

\maketitleabstract

\section{Introduction}
\label{intro}
Understanding biochemical mechanisms such as protein signaling pathways is one of the central pursuits of biomedical research \cite{Martin-90} \cite{arighi2011overview,krallinger2017biocreative,bionlp-2020-sigbiomed}. Biomedical research has advanced tremendously in the past few decades, to the point where we now suffer from ``an embarrassment of riches''. Publications are generated at such a rapid pace (PubMed \footnote{\url{https://pubmed.ncbi.nlm.nih.gov}} 
has indexed more than 1 million publications per year in the past 8 years!) that we need information access applications which can help extract and organize biomedical relations and summarize the biomedical mechanisms underlying them. Developing models that can read biomedical texts and reason about these mechanisms is an important step towards this.

\begin{figure}

    \centering
    \includegraphics[width=\linewidth]{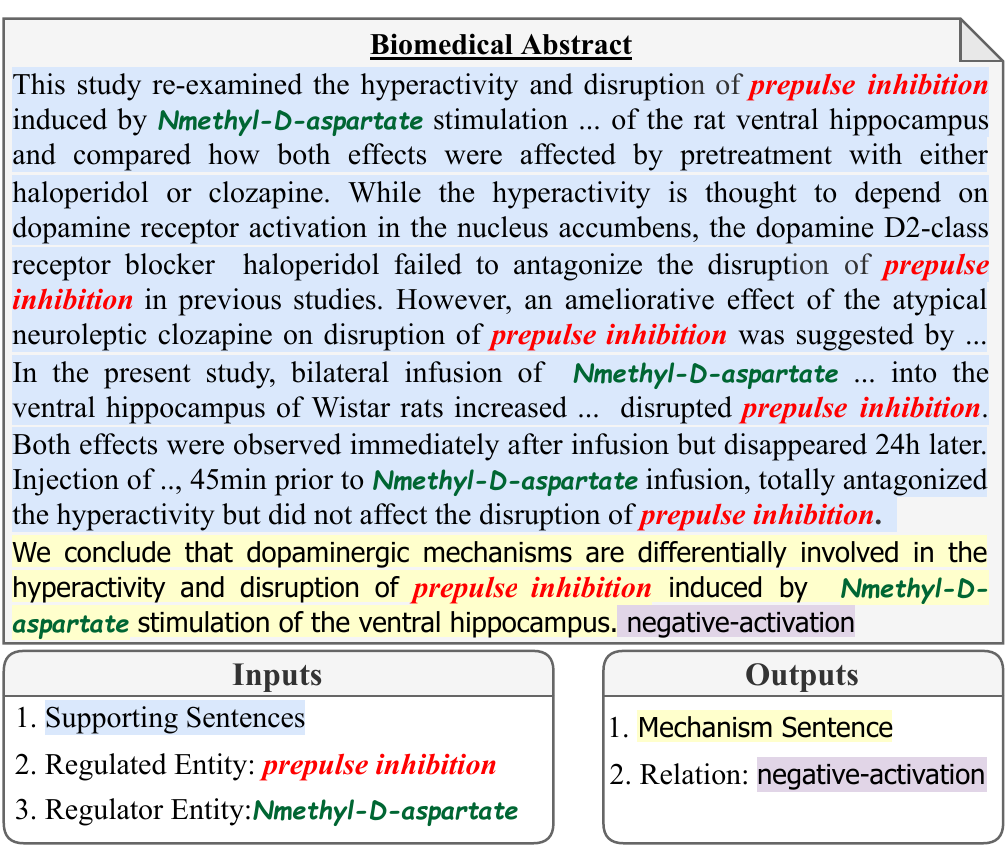}
    \vspace{-1.5em}
    \caption{Biomedical Mechanism Summarization Task: Example of an entry from the \dataset dataset. Some supporting text was removed to save space. The input is the supporting sentences with the main two entities. The output is the relation type and a sentence concluding the mechanism underlying the relationship.}
    \vspace{-1.5em}
    \label{fig:dataexample}
\end{figure}
\begin{figure*}
     \centering
     \small
      \includegraphics[width=\linewidth]{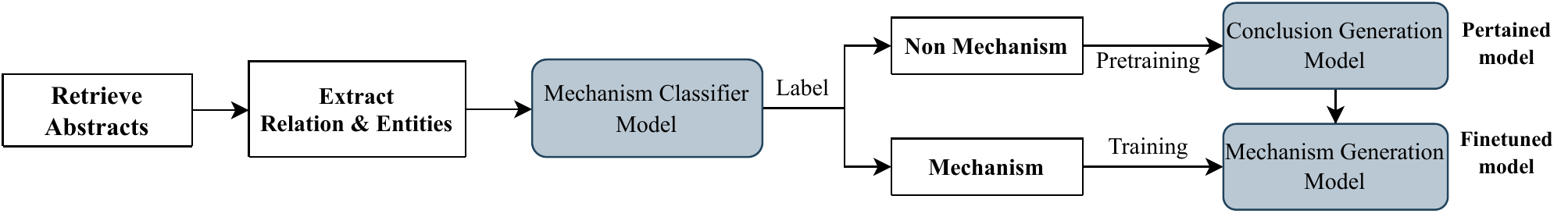}

     \caption{Overview of the semi-automatic bootstrapping process for \dataset creation. We use a mechanism classifier trained with small amount of labeled data to produce weakly-labeled training data for mechanism summarization.}
    
     \label{fig:pipeline_simple}
 \end{figure*} 
In this paper, we introduce a mechanism summarization task which couples text that discusses elements of biomedical mechanisms with their summaries. The task requires models to read text that presents information about the connection between two target entities and generate a summary sentence that explains the underlying mechanism  and the relation between the entities. We see this task from two perspectives. First, summarizing biomedical mechanisms can be seen as part of the broader efforts in extracting~\cite{czarnecki2012text}, organizing~\cite{kemper2010pathtext,kemper2010pathtext,miwa2013method,subramani2015hpiminer,poon2014literome}, and summarizing~\cite{azadani2018graph} biomedical literature that are aimed at providing information access tools for domain experts.

Second, from an NLP perspective this task can be seen as an explainable relation extraction in a biomedical context, where the explanation is the mechanism that provides information about why the relation holds or how it comes about.

A key challenge in addressing such a task lies in creating a large scale dataset necessary for training large neural models. However, building such a dataset manually is a laborious process and requires deep biomedical expertise. To address this, we turn to the structure that exists in biomedical abstracts, make use of related datasets, and devise a  semi-automatic bootstrapping process that builds on a relatively small amount of labeling effort from domain experts.

We introduce \dataset, a large scale dataset that we construct from abstracts of papers that report on biomedical mechanisms. For a given abstract, we create a task instance that consists of a pair of biochemical entities (regulated and regulator), the relationship between them (positive/negative activation), and supporting sentences that provide information about this relationship, and a sentence that summarizes the mechanism underlying the relation (see Figure~\ref{fig:dataexample}). Creating such an instance would require a domain expert to read through an abstract and assess if it contains a biomedical mechanism and locate it if so. This process is difficult to scale.

To address this issue, we introduce a semi-automated annotation process to create a large-scale set for development and automatic evaluation purposes and a clean small-scale manually curated subset of instances for manual evaluation. In particular, the necessary entities and relations are extracted using an existing biomedical information extraction system \cite{Escarcega:2018}. 
To extract mechanism summaries we first collected a small set of mechanism sentences with the help of domain experts. We use this to bootstrap a larger sample by training a mechanism sentence classifier with a biomedical language model (LM) \cite{kanakarajan-etal-2021-bioelectra} and apply it to a large collection of about 611K abstracts that contained a conclusion sentence about the relationship between a pair of entities. The subset that the classifier identifies as containing mechanism sentences is used to create 22K mechanism summarization instances. Five domain experts manually analyzed a dataset sample of 125 instances to construct a clean partition for manual evaluation purposes. The experts also concluded that the generated dataset has reasonable quality, i.e., 84\%. Note that it is common to tolerate some level of noise in the training partitions of automatically constructed NLP datasets. As an example among many, the popular relation extraction dataset by \newcite{yao2010collective} contains over 20\% noise. The overall pipeline is demonstrated in Figure~\ref{fig:pipeline_simple}. In summary, the contributions of this paper are the following:
\begin{itemize}
    \item We introduce the \dataset dataset, the first dataset towards summarizing biomedical mechanisms and the underlying relations between entities. The dataset contains 22K mechanism summarization instances collected semi-automatically, an evaluation partition of 125 instances that were corrected by domain experts. We also create a conclusion generation task from the larger set of 611K abstracts which we use as a pretraining task for mechanism generation models. 
 
\item We benchmark several state-of-the-art language models for the task of generating the underlying biochemical relations and the
corresponding mechanism sentences. We train general domain LMs (GPT2~\cite{radford2019language}, T5~\cite{2020t5}, BART~\cite{lewis2019bart}), as well as science domain adapted versions(scientific GPT2~\cite{papanikolaou2020dare}, and SciFive~\cite{phan2021scifive}) and benchmark their performance through both automatic evaluation and manual evaluation on curated evaluation samples.

 \item  The evaluation by domain experts suggests that this is a high quality dataset coupled with a challenging task, which deserves further investigation.
 
\item To encourage reproducibility and further research, we release the dataset and the code used during its creation. Both are available at \href{https://stonybrooknlp.github.io/SuMe/}{SuMe webpage}.

\end{itemize}

\section{Related Work}
\label{related}

Deep learning models have been widely used in different NLP applications~\cite{gaonkar-etal-2020-modeling,bastan-etal-2020-authors,DBLP:journals/corr/abs-2109-14638,9666618}. Amongst these applications, biomedical NLP is using these models that looks at extracting~\cite{alam2018domain,mulyar2021mt,giorgi2020towards}, organizing~\cite{yuan2020constructing,zhao2020ontosem,lauriola2021learning}, and summarizing information~\cite{DBLP:journals/corr/abs-1804-05685} from scientific literature.

Within this broad context, the mechanism summarization task we introduce broadly relates to previous work in reading and generating information from scientific texts. Most work in this area focus on generating summaries using scientific publication and some times in combination with external information~\cite{yasunaga2019scisummnet,deyoung2020evidence,collins2017supervised}\\
Some works even seek to generate part of the scientific papers. For example,  TLDR~\cite{cachola2020tldr} introduces a task and a dataset to generate TLDRs (Too Long; Didn't Read) for papers. They exploit titles and an auxiliary training signal in their model.  ScisummNet~\cite{yasunaga2019scisummnet} introduces a large manually annotated dataset for generating paper summaries by utilizing their abstracts and citations. TalkSumm~\cite{lev2019talksumm} generates summaries for scientific papers by utilizing videos of talks at scientific conferences.
PaperRobot~\cite{wang2019paperrobot} generates a paper's abstract, title, and conclusion using a knowledge graph. 
FacetSum~\cite{DBLP:journals/corr/abs-1804-05685} used Emerald journal articles to generate 4 different abstractive summaries, each  targeted at specific sections of scientific documents.

In addition to the specifics of the output that we target, our work is different from all these other works because our proposed summarization task is grounded with the underlying biomedical event discussed, rather than focusing on generic summarization, which may lose the connection to the underlying biology that is the core material discussed in these papers. 
We address mechanism generation, 
which can be seen as a combination of explainable relation extraction and summarization. There is a huge body of work that addresses explainable methods (e.g., relation extraction~\cite{shahbazi2020relation} or explainable QA \cite{thayaparan2020survey}). Many prior works in relation and event extraction treat explanations as the task of selecting or ranking sentences that support a relation (e.g., \cite{shahbazi2020relation,ccano2020two,yasunaga2019scisummnet}). Our work differs from these in that it focuses on {\em generating mechanisms} underlying a relation from supporting sentences, rather than identifying existing sentences.

\section{Mechanism Summarization}

\label{task}

Our goal is to develop a task and a dataset that pushes models towards distilling the mechanisms that underlie the relationships between entities from biomedical literature. From a language processing perspective, we can view mechanisms as a form of explanation that justifies the relationship or connection between entities. From a biomedical science perspective, a mechanism provides two types of explanatory information, which we use to characterize mechanism sentences:\\
\textbf{Why is the relation true?} A sentence can be a mechanism, if it explains \textit{why} the relation exists between the two main entities. For example, one protein (say A) might be up-regulate another (say B), which in turn inhibits yet another protein (say C). This provides the causal reasoning to conclude the relation that protein A inhibits protein C. 

\textbf{How does the relation come about?} 
Another kind of explanatory information is the one that describes the process or manner in which the relation exists between the pair of entities. For example, one protein (say A) may activate another protein (say B) via a specific process. 
\vspace{0.3em}

These provide a way to specify what constitutes a mechanism sentence and help us to locate mechanism sentences in the literature. In particular, we consider abstracts which discuss studies that lead to conclusions about such mechanisms. Typically, these abstracts provide a short set of sentences that describe the goals of the study, the methods used, the experimental observations, the findings, which can be used to substantiate the conclusions that establish the relation of interest, and the mechanism underlying the relation. This   suggests a language processing task that tests for ability to understand biomedical mechanisms: given the preceding sentences in the abstract can a model accurately generate the underlying mechanism?

\subsection{Task Definition}
Given a set of sentences from a scientific abstract (referred to as {\em supporting sentences}) and a pair of entities $(e_i, e_j)$ that are the focus of the abstract (referred to as {\em focus entities}), generate the \textit{conclusion sentence} that explains the mechanism behind the pair entities and output a relation that connects these entities (e.g., positive\_activation$(e_i, e_j)$). 
Figure~\ref{fig:dataexample} shows an example of such a tuple of supporting sentences, focus entities, relation, and mechanism sentence. 
As the example illustrates, mechanism sentences describe some pathway often involving another entity or a process (e.g., {\em dopaminergic mechanism}), require identifying and combining information from multiple relevant sentences, and non-trivial inferences  regarding the relationship between the entities (e.g., recognizing that the different effects on \emph{prepulse inhibition} imply differential involvement).  

The task definition suggests what we need to build a dataset. Given an abstract of a scientific literature we need four pieces of information: 
(1) the two focus entities of the abstract; (2) the relation between entities; (3) sentences from the abstract in support of this relation; and (4) the conclusion sentence where the mechanism underlying the relation is summarized.

\section{\dataset Dataset}
\label{data}
 
We aim to create a large scale dataset for the mechanism summarization task defined above. However, identifying instances for this task requires domain expertise and cannot be easily done at scale. Instead, here we employ a bootstrapping process, where we first annotate a small amount of data to build a mechanism sentence classifier that can then helps us collect a large scale dataset for mechanism summarization. The key observation here is that identifying sentences that express a mechanism is a simpler task than the targeted mechanism summarization task, and, thus, should be learnable from smaller amounts of data. We outline the process we use for creating our mechanism summarization dataset, \dataset, and an expert evaluation of its quality next. 

\subsection{\dataset Construction Process}

\begin{figure*}[ht!]
     \centering
     \small
     \includegraphics[scale=0.75]{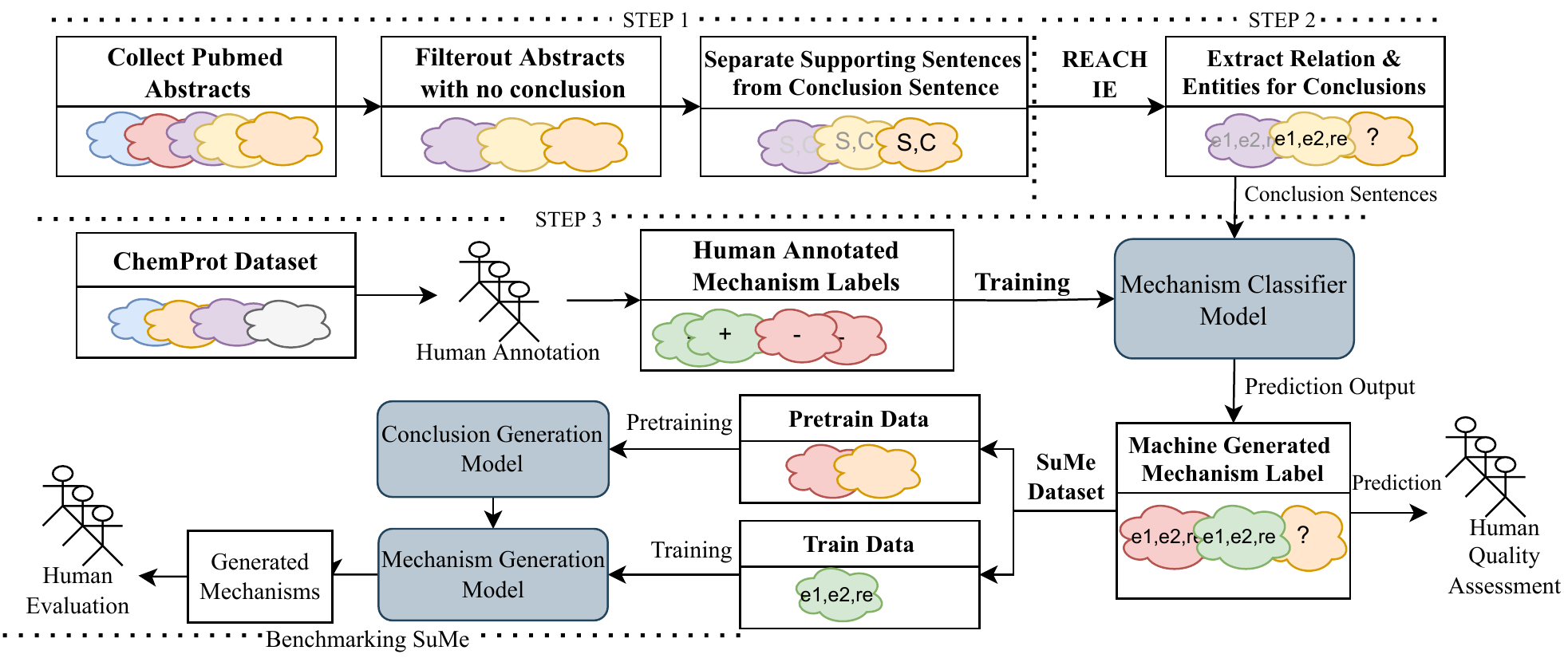}

    \caption{The bootstrapping pipeline for \dataset collection and human evaluation. The main idea behind the pipeline is to collect relatively easy to acquire judgments from domain experts to then bootstrap and generate a weakly-labeled large training corpus. We further assess the quality of the resulting dataset through another round of human evaluation, which also yields a smaller curated evaluation dataset.}
    \vspace{-1em}
     \label{fig:pipeline}
 \end{figure*}

We construct \dataset using biomedical abstracts from the PubMed open access subset\footnote{\url{https://pubmed.ncbi.nlm.nih.gov}}. 
Starting from 1.1M scientific papers~\footnote{We used all papers available in  \href{https://ftp.ncbi.nlm.nih.gov/pubmed/baseline/}{NIH active directory}}, we use the following sequence of bootstrapping steps to construct \dataset. The following steps are also elaborated in Figure~\ref{fig:pipeline}. 

\noindent\textbf{1. Finding Conclusion Sentences:} First, we use simple lexical patterns to find abstracts with a clearly specified conclusion sentence. All abstracts which has any form of \textit{conclude} word (\textit{conclusion, concluded, concluding, concludes}, etc.) at the very end of the text are extracted here. We use this matching process to also split the abstracts into the set of supporting sentences (the ones that lead up to the conclusion) and one conclusion sentence (the one that includes the \textit{conclude} word). 

\noindent\textbf{2. Extracting Main Entities \& Relation}
Starting with the abstracts which are now in the form of (supporting sentences, conclusion sentence), we then run a biomedical relation extractor, REACH~\cite{Escarcega:2018}, which can identify protein-protein and chemical-protein relations between entities. 
In this work, we focus on the relations where one entity is the controller and another entity is the controlled entity and the relation between them is either \textit{positive/negative activation} or \textit{ positive/negative regulation}. If an abstract does not contain any such relation, we keep it for the pretraining step (as described in Section~\ref{DA}); otherwise we use it for the main task.

\noindent\textbf{3. Filtering for Mechanism Sentences:} 

We then filter out the instances to only retain those whose conclusion sentences are indeed a mechanism sentence. To this end, we devised a bootstrapping process where we first collect supervised data to train a  classifier. To collect likely mechanism sentences we made use of the ChemProt~\cite{peng2019transfer} relation extraction dataset which contains sentences annotated with positive and negative regulation relations between entities. However, not all of these sentences necessarily explain the mechanism behind these relations. We asked 21 experts (grad students in a biomedical department) to inspect each sentence and rate whether it explains the mechanism behind the ChemProt annotated relation on a four-point Likert scale. For each sentence, an annotator can select between \textit{Clearly a Mechanism, Plausibly a Mechanism, Clearly not a Mechanism,} and \textit{Not Sure.} Each sentence is annotated by three experts and we find the inter-annotator agreement between users to be $\kappa=73\%$ (Fleiss Kappa~\cite{landis1977measurement}). The final label for a sentence is selected based on the majority voting after combining \textit{Clearly a Mechanism} and \textit{Plausible a Mechanism} labels. Finally, each sentence is labeled as a \textit{Mechanism}, or \textit{Non-Mechanism}. The resulting dataset contained 439 \textit{Mechanism} sentences \edit{(264 \textit{Clearly}, 175 \textit{Plausibly})} and 447 \textit{Non-Mechanism} sentences. 

Using this small scale mechanism sentence dataset, we train binary classifiers to identify mechanism sentences, where the positive label indicates that the underlying sentence is a mechanism sentence. We fine-tuned multiple transformer-based models: BioBERT~\cite{lee2020biobert}, SciBERT~\cite{Beltagy2019SciBERT}, BiomedNLP~\cite{pubmedbert}, and BioELECTRA~\cite{kanakarajan-etal-2021-bioelectra} models. Each model is fitted with a non-linear classification layer that takes the output representation for the [CLS] token. The classification layer and top three layers of the transformer are finetuned using the annotated data\footnote{All models used are base  versions with 768 hidden size and 12 layers. We set the learning rate to be $2e-4$ with a decay of $0.001$}. \edit{We used 80\%-20\% split for train-test.} BioELECTRA performed the best with 74\% macro F1 for mechanism sentence classification. 

We use this trained mechanism sentence classifier to label all conclusion sentences from the previous step and instances which are predicted to be mechanism sentences are used to create the mechanism generation of the \dataset dataset.

We separate out the abstracts which are predicted to have non-mechanism sentences as additional data. We can define a broader conclusion generation task, which can be be used as a pre-training task for the generative models that eventually use for the mechanism summarization task (as  we describe in Section~\ref{DA}). 

The above procedure results in a dataset that allows us to define the following mechanism summarization task: Given a set of supporting sentences from an abstract and a pair of entities $(e_i, e_j)$, generate a relation that connects these entities and a sentence that explains the mechanism that was the focus of the study. The statistics of the dataset are shown in Table~\ref{tab:data_stats}. The dataset consists of three subsets, the training set with about 20k instances which the parameters of the model are trained with, the validation set (Dev) for tuning hyper parameters and choosing the best model, and the test set which is not used until the final evaluation. There is also a small set of 125 instances which is curated by experts and is used as another test set but is not reported in this table.

\begin{table}
    \centering
    \begin{tabular}{c|c|c|c} 
        Dataset &Train & Dev & Test  \\ \hline \hline
        Abstracts &20765&1000&1000\\
        Avg. \#words in conc.&33.7&34.9&33.5 \\
        Avg. \#words in supp.&187.5&187.9&186.7 \\
        Avg. \#sent. in supp. &12.15&12.44&12.33\\
        \#Unique controller &8094&759&777\\
        \#Unique controlled &6684&717&687\\
        \#Unique pair entities &19229&988&989\\
        \#Unique entities   &12685&1357&1364\\
    \end{tabular}
    \caption{Dataset Statistics: Each dataset contains a number of unique abstracts, a supporting set (supp.), a mechanism sentence (conc.) a pair of entities. The first entity is called the regulator entity (regulator) and the second one is called the regulated entity (regulated)}
    \label{tab:data_stats}
\end{table}

\subsection{\dataset Quality}
Our process creates a large scale, albeit, a bootstrapped dataset that can be used to train large language generation models. What is the quality of this dataset? To assess this we asked five biomedical experts to evaluate a random sample of 125 sentences from the dataset. The experts were given the set of input supporting sentences, the potential  mechanism sentence, and the relation between main entities. Our aim is two fold, first to evaluate the quality of the data collection process, second to collect a clean human evaluated dataset which can be used as an extra test set. 

The experts were asked to assess errors in the relation label, mechanism, and the need for background knowledge:
\begin{enumerate}[noitemsep]
\vspace{-0.3em}
    \item \textbf{Relation Errors:} Is the expected output relation associated with the instance valid?     
    \item \textbf{Mechanism vs. Non-mechanism:} Is the output sentence expected for this sentence an actual mechanism sentence? 
    \item \textbf{Background Knowledge: } Can the information in the output sentence concluded from the information in the input supporting sentences?
\end{enumerate}

The results of the dataset evaluation are shown in Table~\ref{tab:qual}. Only 16\% of the data has some error either from relation extraction (question 1) or contains a non-mechanism output sentence (question 2).This evaluation shows that the generated dataset is of reasonable 
quality, and can serve as a meaningful resource for training models for mechanism summarization. The clean subset that has no relation or mechanism errors is used as an extra test for evaluation. Last, the experts also rated 15\% of the instances to require background knowledge (question 3)indicating the fraction of hard instances.

\begin{table}
    \centering
    \begin{tabular}{c|c} 
         Quality & Correct \\ \hline \hline
         Entities \& Relation Extraction &  90\% \\
         Mechanism Sentence Classifier &  85\% \\
         \hline
         Instances w/o noise & 84\%

    \end{tabular}
    \caption{Dataset Quality: We asked three main questions. This table shows what percentage of each category is acceptable. The last question shows what percentage of the sentences are approved in all questions.}
    \vspace{-1.2em}
    \label{tab:qual}
\end{table}

 \section{Evaluation}
 \begin{table*}[ht!]
    \centering

    \begin{tabular}{c||c|c|c|c|c}
         Model&   RG (F1) &BLEURT & Rouge-1 &Rouge-2 &Rouge-L \\ \hline \hline
         BART &76&42.49&46.54&25.92&35.34\\
         GPT2&74&44.19&46.54&28.32&38.78 \\
         T5 &72&44.41&48.26&27.63&38.77 \\
         \hline\hline
         GPT2-Pubmed &78&46.33&48.37&29.55&40.19 \\
         SciFive&\textbf{79}&\textbf{47.81}&\textbf{52.10}&\textbf{32.62}&\textbf{43.31} \\

    \end{tabular}
    \caption{Benchmarking performance of strong language generation models and some domain-adapted models. We present standard automatic evaluations measures for the mechanism sentence generation task along with F1 for the generated relations. The science domain versions of both GPT2 and T5 work better than the original versions.}
    \vspace{-1em}
    \label{tab:benchmarks}
\end{table*}
\label{exper}
Our evaluation focuses on the following questions:
\begin{enumerate}[noitemsep,nolistsep]
    \item Benchmarking: What is the performance of generic and domain-adapted large scale language generation models on \dataset?
    \item Effect of pretraining: What is the impact of using the additional data via pretraining?
    \item Effect of modeling supporting sentences: What is the impact of selecting a subset of supporting sentences?
    \item Error analysis: What are the main failure modes of language generation models?
    
\end{enumerate}

\subsection{Experimental Setup} 

We use \dataset to benchmark language generation models and measure their ability to correctly identify the relation between the focus entities and to summarize the mechanism behind the relation based on the input sentences from the abstract. 

\noindent{\textbf{Models:}} We compare pretrained GPT-2~\cite{radford2019language}, T5~\cite{raffel2020exploring}, BART~\cite{lewis2019bart} models and two domain-adapted models, GPT2-Pubmed~\cite{papanikolaou2020dare}, and SciFive~\cite{phan2021scifive}, which were trained on scientific literature.

\noindent{\textbf{Evaluation Metrics:}}
We conduct both automatic and manual evaluation of the model outputs.

\noindent\textit{Relation Generation (RG):} The models are supposed to first generate the relation type (positive or negative regulation) and then generate the mechanism that underlies this relation. We evaluate the model's output as we would for a corresponding classification task, i.e., the generated relation is deemed correct if it exactly matches the correct relation name. We report F1 numbers for this binary classification task. 
   
\noindent\textit{Mechanism Generation:} We evaluate the quality of the generated explanations using two language generation metrics: the widely-used ROUGE~\cite{lin2004rouge} scores that rely on lexical overlap, and BLEURT scores~\cite{sellam2020bleurt} which aim to capture semantic
similarity between the generated and the gold reference.

We use a recent version, the BLEURT-20 model that has been shown to be more effective~\cite{Pu2021LearningCM} . We compare the generated text as the hypothesis against the actual text as the reference. 

\noindent{\textbf{Fine-tuning and Training Details:}} All models are original base models published by HuggingFace that were fine-tuned on the training portion of \dataset for 20 epochs. For each model, we evaluate the average of  BLEURT and Rouge-L score on the validation (Dev) set and the one with the highest average is chosen for prediction. The learning rate is set to 6e-5, we use AdamW~\cite{DBLP:journals/corr/abs-1711-05101} optimizer with $\epsilon=1e-8$. The input token is limited to 512 tokens, and the generated token is maxed out at 128. We select batch size of 8 with gradient accumulation steps of two. 

\label{BM}

\subsection{Automatic Evaluation Results}

Table~\ref{tab:benchmarks} compares the performance of the five language generation models on both the relation generation (RG) and mechanism generation tasks. 

The domain-adapted models, GPT2-Pubmed and SciFive, fare better than fine-tuning the standard pre-trained models for both relation and mechanism generation tasks. SciFive achieves the best performance with more than a $7.5\%$ increase in BLEURT score and more than $9.7\%$ increase in RG F1 over the standard T5 model, highlighting the importance of domain adaptation for the \dataset tasks defined over scientific literature. 

The overall numbers (coupled with the human evaluation in Section~\ref{sec:manualeval}) suggest that mechanism generation is a difficult and challenging task. 

The models achieve better performance on the relation generation task but there is still a substantial room for improvement here with the best model achieving an F1 of 79. If the model is unable to generate the relation correctly, then the mechanism it generates is not useful. Ideally we want models to correctly generate both the relation and the mechanism that underlies it. We also evaluated the correlation between BLEURT score and relation generation classification score. Our analysis shows that when the model generates an accurate relation, it gets higher BLEURT score while when it generates an incorrect relation, its gets a 10\% lower BLEURT score (50.02 vs 45.08)

\subsection{Pretraining with Conclusion Generation}
\label{DA}
Next we analyze the impact of pre-training the models on the related task of generating conclusion (instead of mechanism) sentences, for which we can obtain data at scale without any manual labeling effort. We collected all abstracts from PubMed that ended with a conclusion sentence. We can create training instances on these abstracts in the same format as we did for the mechanism generation instances. The only difference here is the output sentences are conclusion sentences and not necessarily mechanisms. We call this the conclusion generation task. \dataset includes 611K instances of this kind which is an order of magnitude larger than the mechanism summarization instances and can be seen as a form of data augmentation.

We study the effect of this pretraining task by varying the amount of pretraining data. We analyze the impact in terms of the overall effectiveness and the amount of fine-tuning (number of epochs) needed to converge when finetuning. 

\noindent\textbf{Pretraining Data Size:}
\begin{figure}
    \centering
    
    \includegraphics[width=0.95\linewidth]{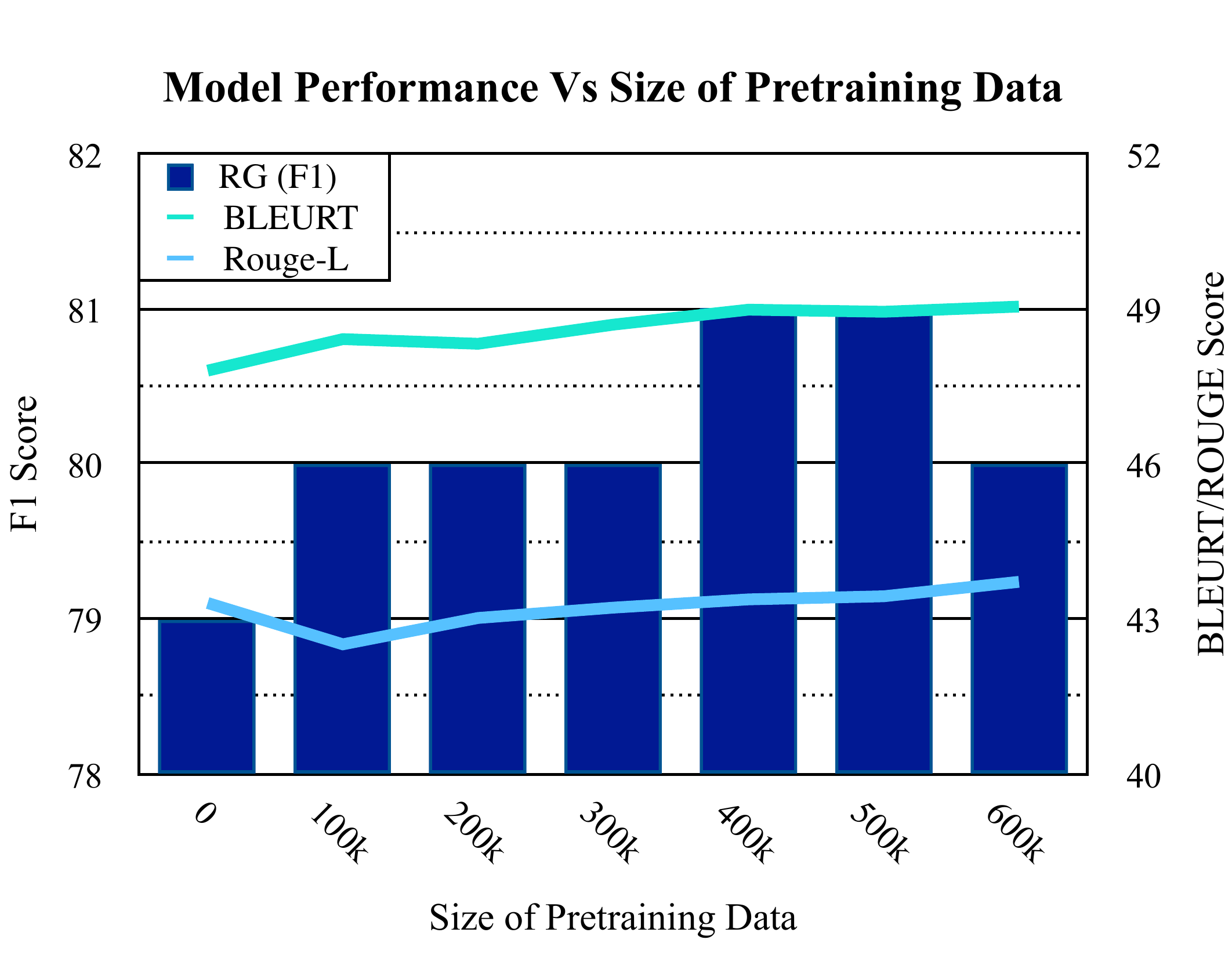}
      \vspace{-0.9em}
    \caption{Comparison of relation generation (RG) F1 (left y-axis/blue bars) and the mechanism generation measures (right y-axis/teal+blue curves) against the amount of pretraining. As we increase the size of the pretraining data, the model performance improves.}
    \vspace{-1.2em}
   
    \label{fig:pretrain}
  
\end{figure}
We pretrain the SciFive model on the conclusion generation task with increasing amount of data (100K increments), and measure the performance of finetuning the pretrained models on the mechanism summarization task. Figure~\ref{fig:pretrain} shows that performance increases with more data available for pretraining, suggesting that pretraining is beneficial for learning to generate mechanisms.

\begin{figure}
    \centering
    \includegraphics[width=0.95\linewidth]{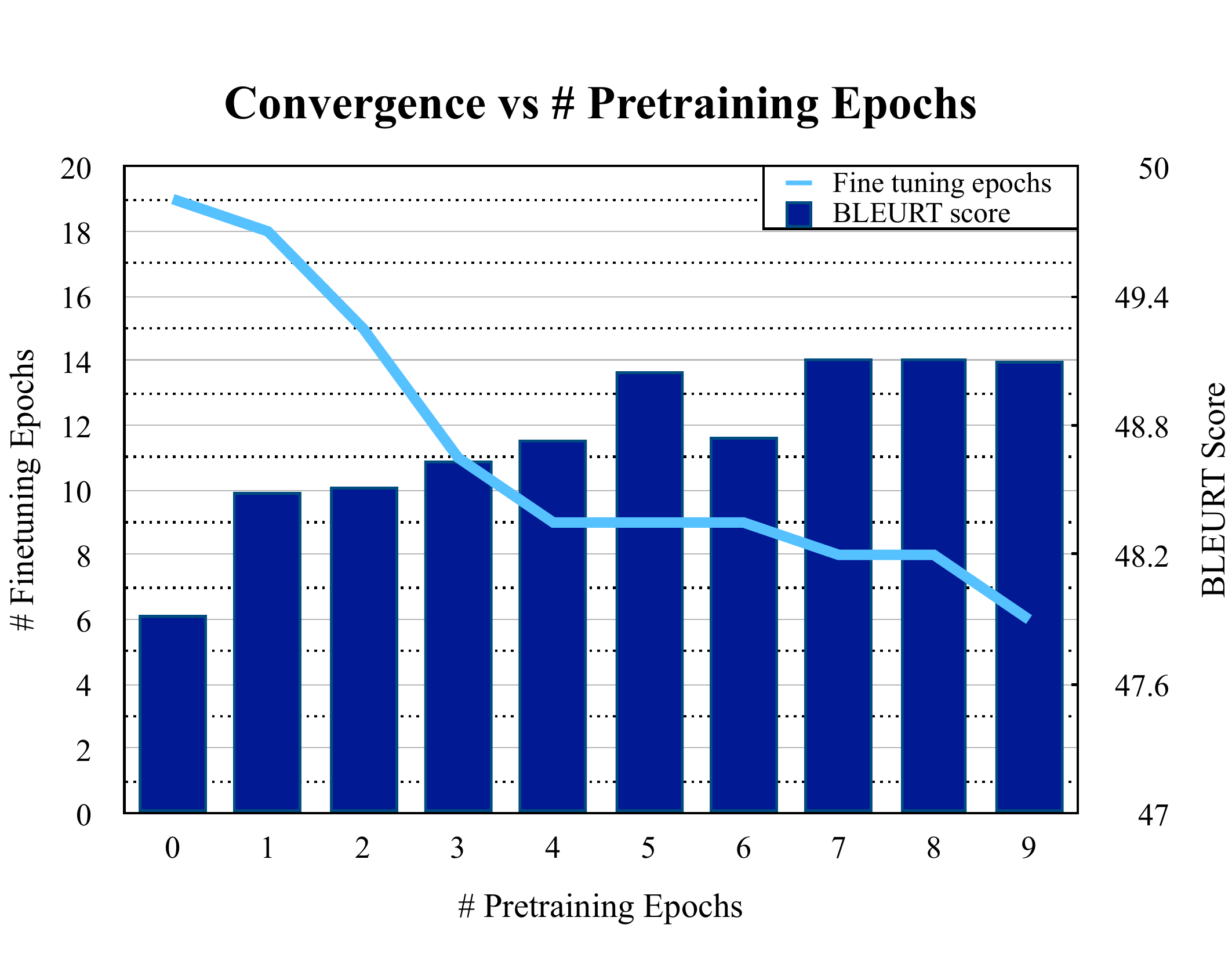}
 \vspace{-0.6em}
    \caption{Number of pretraining epochs vs. fine-tuning epochs for each pretrained model until convergence.}
    \vspace{-1em}
    \label{fig:pretrain_ep}
\end{figure}

\noindent\textbf{Number of Epochs:}
We also compare the impact of the amount of pretraining on the number of epochs needed for convergence in fine-tuning. Figure~\ref{fig:pretrain_ep} compares pretrained models with different number of pretraining epochs (x-axis) in terms of their overall effectiveness (BLEURT score bars) and the number of epochs to convergence (Finetuning epochs curve). The figure shows that when we continue pretraining, not only does the resulting model performs better, but it also converges sooner taking fewer number of epochs to reach higher effectiveness. Together these results suggest potential for the auxiliary data available in the \dataset dataset.

\subsection{Modeling Supporting Sentences}
Will it help to model the subset of sentences within the inputs sentences that provide the best support for generating the mechanism sentence? This kind of an extractive step has been used previously in summarization tasks to reduce the amount of irrelevant information in the input~\cite{narayan2018don,liu2019text}. To understand the utility of this, we built a pseudo-oracle that finds the sentences that have the best overlap (measured via BLEURT score~\cite{sellam2020bleurt}) with the output mechanism sentence. Then, we trained the SciFive model and pretrained version to only use the top few sentences according to BLEURT score such that input size is now half of the original input size. Using this subset instead of the entire subset provides BLEURT score improvements only for the basic SciFive model and the gains reduce when we use the pretrained model.  Unlike standard summarization tasks there are fewer completely unrelated sentences in the abstracts and generating the mechanism sentences remains challenging even when we are able to identify the most relevant sentences within this set. This suggests that the task remains hard even when the most important sentences are somehow known to the model.


\begin{table}
    \centering
   
    \begin{tabular}{c||c|c}
        Supporting Set &  BLEURT &Rouge-L \\ \hline \hline
        SciFive & 47.81&43.31\\
         +Oracle & 49&43.07\\
         +Pretraining &49.05&43.72\\
         +Pretraining+Oracle &49.64&43.81\\
        

    \end{tabular}
    \caption{The effect of selecting supporting sentences with highest BLEURT score. }
    \vspace{-1.1em}
 
    \label{tab:pretrain_ep}
\end{table}

\subsection{Manual Evaluation}
\label{sec:manualeval}

We also conduct a manual evaluation of the outputs from the best model --- the SciFive model that was pretrained with the conclusion generation task. We asked three biomedical experts to evaluate output sentences for 100 instances and answer three questions \edit{(It took $\sim5$ minutes per expert per instance)}:
\begin{enumerate}[noitemsep,nolistsep]
    \item Does the generated sentence contain a mechanism that explains the relation between the two main entities?
    \item Is the information in the generated mechanism supported by the information contained in the supporting set of sentences?
    \item Is the generated mechanism factually correct? 
    
\end{enumerate}

The results show that the model learns to generate some mechanism sentence that connects the main entities for most of the instances (79\%). It also produces output for which there is support in the input sentences in a majority of the cases (53\%). Last, experts found that the output statements to be scientifically correct in many cases (58\%). In summary, however, only 32\% of the outputs were acceptable for all questions, and were deemed to be good mechanism sentences. This again highlights the significant challenge posed by this task.

\subsection{Error Analysis}
\begin{table*}[ht!]
    \centering
    \begin{tabular}{p{0.52\linewidth} | p{0.43\linewidth}} 
      \textbf{Gold Mechanism}  & \textbf{Generated Mechanism}  \\ \hline \hline
      On the basis of these observations, we conclude that \textit{IL-18} induces  \textit{MCP-1} production   \textcolor{blue}{through the PI3K/Akt and MEK/ERK1/2 pathways in macrophages.} & In conclusion, \textit{IL-18 } induces \textit{ MCP-1} production \textcolor[rgb]{0,0.3,0}{in macrophages through the PI3K/Akt and MEK/ERK1/2 pathways.}\\ \hline
      Taken together, we conclude that  \textit{DeltaNp73}  negatively regulates \textit{ NGF-mediated neuronal differentiation}  \textcolor{blue}{by transrepressing TrkA.} & In conclusion, our results indicate that \textit{DeltaNp73 } negatively regulates \textit{NGF-mediated neuronal differentiation} \textcolor[rgb]{0,0.3,0}{by transcriptionally repressing the expression of TrkA.} \\ \hline
      In conclusion, the increase in  \textit{SNGFR}  caused by \textit{ANF}  is associated with \textcolor{blue}{an increase in glomerular capillary hydraulic pressure and with a blunted maximal tubuloglomerular feedback response.} & We conclude that \textit{ANF} inhibits \textcolor[rgb]{0,0.3,0}{maximal tubuloglomerular feedback by increasing glomerular capillary hydraulic pressure} and by decreasing \textit{ SNGFR}. \\ \hline 
      We conclude that, \textcolor[rgb]{0,0,1}{without modulatory factors which play a role in vivo}, \textit{ NGF}  can enhance the  synthesis of \textit{tyrosine} hydroxylase n sympathetic ganglia in vitro, provided organ culture conditions which permit optimal survival of adrenergic neurons are selected. &This conclusion was further supported by pulse-labeling of \textit{tyrosine} hydroxylase with [3H]leucine, which showed that \textit{NGF } increased  synthesis  of \textit{tyrosine} \textcolor[rgb]{1,0,0}{in sympathetic ganglia by approximately 50\%}.  \\ \hline
      We conclude that \textcolor[rgb]{0,0,1}{20 mm alcohol/submaximal CCK as well supramaximal CCK stimulation can trigger pathologic basolateral exocytosis in pancreatic acinar cells }via PKC alpha-mediated activation of  \textit{Munc18c} , which enables  \textit{Syntaxin-4} to become receptive in forming a SNARE complex in the BPM. & We conclude that alcohol can induce a clinically relevant form of pancreatitis  \textcolor[rgb]{1,0,0}{by blocking apical exocytosis and redirecting exocytosis to less efficient BPM, mimicking supramaximal CCK stimulation.} \\ \hline
      We conclude that \textcolor[rgb]{0,0,1}{in the presence of high doses of insulin}, \textit{FSH }  decreases  \textit{aromatase} activity, and an uncoupling of P450 \textit{ aromatase}  mRNA and  \textit{aromatase} activity occurs. & In conclusion, insulin stimulates \textit {aromatase} activity in bovine granulosa cells at low doses but \textcolor[rgb]{1,0,0} {fails to stimulate activity at higher doses of insulin.} 
    \end{tabular}
    
    \caption{Examples of the generated outputs by the model. The first three are good outputs where the mechanism is a simple paraphrase of the expected gold mechanism, while the next three illustrate the types of semantic errors we observe. The main entities are makred in \textit{Italics}. The phrase explaining the mechanism in gold data is in \textcolor[rgb]{0,0,1}{blue}, in good generation is in \textcolor[rgb]{0,0.3,0}{green}, and in bad generation is in \textcolor[rgb]{1,0,0}{red}. }
    \vspace{-1.5em}
    \label{tab:output}
\end{table*}
To understand the frequent failure modes of the model, we manually categorized the errors in 100 outputs that had the \textbf{worst BLEURT}  scores with the reference mechanism sentences.

\begin{enumerate}[nolistsep, noitemsep]
\item \textbf{Missing Entities (35\%)} -- The most prevalent issue is the absence of one of the main entities in the generated sentence. Despite this being a necessary feature in all of the mechanism sentences in the training data, the prevalence of this error shows that models find it difficult to track the main entities during generation.

\item \textbf{Incorrect Mechanism (24\%)} -- The model is unable to generate the correct mechanism even though it is able to identify the correct relation and fills in some information that is either unrelated to or unsupported by the input sentences. 

\item \textbf{Flipped Relation (19\%)} --  The model predicts the incorrect relation and generates a mechanism that is faithful to this incorrect relation. Improving relation generation is thus an important step for improving mechanism generation.

\item \textbf{Non Mechanisms (11\%)} -- While the model learns to generate mechanism like sentences for the most part, it sometimes still fails to produce sentences that contain any mechanism at all. 

\item \textbf{Multiple pieces of information (11\%)} -- Some complex mechanisms require combining bits of information from different input sentences. The model generates only a part of such mechanisms.
\end{enumerate}

\subsection{Word Analysis}
We further analyzed the unigrams of the supporting sentences corresponding to the instances where the model was most confident in its generated mechanism and where it was least confident. The analysis shows that when the words \textit{'binding', 'caused', 'demonstrated', 'dose dependent', 'investigated', 'result',} and \textit{'performed'} are available in the supporting sentences the model can generate explanation sentences with higher quality. This shows that when the supporting sentences convey causal relation and reasoning the model is most confident about generating mechanisms.

Table~\ref{tab:output} shows example generated mechanisms. The first three showcase good outputs whereas the next three are examples of incorrect ones. In the good ones, the first is a generated mechanism that is almost identical to the gold mechanism with only a slight syntactic change. The second is a generated mechanism which also conveys the gold mechanism accurately but with a paraphrasing that expands the technical term \textsc{transpressing}. In the last three examples with incorrect information, the first shows a bad output which contains a mechanism but not of the relation connecting the main entities. The next is a case where the information is correct but it does not even mention the main entities. The last one is an example one of the entities are missing (\textit{FSH}) and the generated text is about another relation. 

\section{Conclusions}
We introduced \dataset, a dataset for biomedical mechanism summarization. This dataset is coupled with a challenging summarization task, which requires the generation of the relation between main entities as well as a textual summary of the mechanism which explains the reason behind the underlying relation. This dataset is collected using the sentences from actual publication abstracts. We also introduce an easier and scalable pretraining task which improves the baselines by augmenting a larger set of sentences to the main dataset.  We evaluated the complexity of the task using multiple state-of-the-art transformer based models. Our evaluation suggests that the proposed task is learnable, but we are far from solving it. The expert analysis also suggests the difficulty and importance of the task.

All in all, we believe that \dataset dataset and associated task are a useful step towards building true information-access applications for the biomedical literature. 

\subsection*{Acknowledgments}
This work was supported in part by the National Science Foundation under grants  IIS-1815358 and IIS-1815948.

\section{Bibliographical References}\label{reference}
\bibliographystyle{lrec2022-bib}
\bibliography{lrec2022}

\begin{thebibliography}{}

\bibitem[\protect\citename{Alam \bgroup et al.\egroup }2018]{alam2018domain}
Alam, F., Joty, S., and Imran, M.
\newblock (2018).
\newblock Domain adaptation with adversarial training and graph embeddings.
\newblock {\em arXiv preprint arXiv:1805.05151}.

\bibitem[\protect\citename{Arighi \bgroup et al.\egroup
  }2011]{arighi2011overview}
Arighi, C.~N., Lu, Z., Krallinger, M., Cohen, K.~B., Wilbur, W.~J., Valencia,
  A., Hirschman, L., and Wu, C.~H.
\newblock (2011).
\newblock Overview of the biocreative iii workshop.
\newblock {\em BMC bioinformatics}, 12(8):1--9.

\bibitem[\protect\citename{Azadani \bgroup et al.\egroup
  }2018]{azadani2018graph}
Azadani, M.~N., Ghadiri, N., and Davoodijam, E.
\newblock (2018).
\newblock Graph-based biomedical text summarization: An itemset mining and
  sentence clustering approach.
\newblock {\em Journal of biomedical informatics}, 84:42--58.

\bibitem[\protect\citename{Bastan \bgroup et al.\egroup
  }2020]{bastan-etal-2020-authors}
Bastan, M., Koupaee, M., Son, Y., Sicoli, R., and Balasubramanian, N.
\newblock (2020).
\newblock Author{'}s sentiment prediction.
\newblock In {\em Proceedings of the 28th International Conference on
  Computational Linguistics}, pages 604--615, Barcelona, Spain (Online),
  December. International Committee on Computational Linguistics.

\bibitem[\protect\citename{Beltagy \bgroup et al.\egroup
  }2019]{Beltagy2019SciBERT}
Beltagy, I., Lo, K., and Cohan, A.
\newblock (2019).
\newblock Scibert: Pretrained language model for scientific text.
\newblock In {\em EMNLP}.

\bibitem[\protect\citename{Cachola \bgroup et al.\egroup
  }2020]{cachola2020tldr}
Cachola, I., Lo, K., Cohan, A., and Weld, D.~S.
\newblock (2020).
\newblock Tldr: Extreme summarization of scientific documents.
\newblock {\em arXiv preprint arXiv:2004.15011}.

\bibitem[\protect\citename{{\c{C}}ano and Bojar}2020]{ccano2020two}
{\c{C}}ano, E. and Bojar, O.
\newblock (2020).
\newblock Two huge title and keyword generation corpora of research articles.
\newblock {\em arXiv preprint arXiv:2002.04689}.

\bibitem[\protect\citename{Cohan \bgroup et al.\egroup
  }2018]{DBLP:journals/corr/abs-1804-05685}
Cohan, A., Dernoncourt, F., Kim, D.~S., Bui, T., Kim, S., Chang, W., and
  Goharian, N.
\newblock (2018).
\newblock A discourse-aware attention model for abstractive summarization of
  long documents.
\newblock {\em CoRR}, abs/1804.05685.

\bibitem[\protect\citename{Collins \bgroup et al.\egroup
  }2017]{collins2017supervised}
Collins, E., Augenstein, I., and Riedel, S.
\newblock (2017).
\newblock A supervised approach to extractive summarisation of scientific
  papers.
\newblock {\em arXiv preprint arXiv:1706.03946}.

\bibitem[\protect\citename{Czarnecki \bgroup et al.\egroup
  }2012]{czarnecki2012text}
Czarnecki, J., Nobeli, I., Smith, A.~M., and Shepherd, A.~J.
\newblock (2012).
\newblock A text-mining system for extracting metabolic reactions from
  full-text articles.
\newblock {\em BMC bioinformatics}, 13(1):1--14.

\bibitem[\protect\citename{Demner-Fushman \bgroup et al.\egroup
  }2020]{bionlp-2020-sigbiomed}
Dina Demner-Fushman, et~al., editors.
\newblock (2020).
\newblock {\em Proceedings of the 19th SIGBioMed Workshop on Biomedical
  Language Processing}, Online, July. Association for Computational
  Linguistics.

\bibitem[\protect\citename{DeYoung \bgroup et al.\egroup
  }2020]{deyoung2020evidence}
DeYoung, J., Lehman, E., Nye, B., Marshall, I.~J., and Wallace, B.~C.
\newblock (2020).
\newblock Evidence inference 2.0: More data, better models.
\newblock {\em arXiv preprint arXiv:2005.04177}.

\bibitem[\protect\citename{Gaonkar \bgroup et al.\egroup
  }2020]{gaonkar-etal-2020-modeling}
Gaonkar, R., Kwon, H., Bastan, M., Balasubramanian, N., and Chambers, N.
\newblock (2020).
\newblock Modeling label semantics for predicting emotional reactions.
\newblock In {\em Proceedings of the 58th Annual Meeting of the Association for
  Computational Linguistics}, pages 4687--4692, Online, July. Association for
  Computational Linguistics.

\bibitem[\protect\citename{Giorgi and Bader}2020]{giorgi2020towards}
Giorgi, J.~M. and Bader, G.~D.
\newblock (2020).
\newblock Towards reliable named entity recognition in the biomedical domain.
\newblock {\em Bioinformatics}, 36(1):280--286.

\bibitem[\protect\citename{Gu \bgroup et al.\egroup }2020]{pubmedbert}
Gu, Y., Tinn, R., Cheng, H., Lucas, M., Usuyama, N., Liu, X., Naumann, T., Gao,
  J., and Poon, H.
\newblock (2020).
\newblock Domain-specific language model pretraining for biomedical natural
  language processing.

\bibitem[\protect\citename{Heidari \bgroup et al.\egroup }2021]{9666618}
Heidari, M., Zad, S., Hajibabaee, P., Malekzadeh, M., HekmatiAthar, S., Uzuner,
  O., and Jones, J.~H.
\newblock (2021).
\newblock Bert model for fake news detection based on social bot activities in
  the covid-19 pandemic.
\newblock In {\em 2021 IEEE 12th Annual Ubiquitous Computing, Electronics
  Mobile Communication Conference (UEMCON)}, pages 0103--0109.

\bibitem[\protect\citename{Kanakarajan \bgroup et al.\egroup
  }2021]{kanakarajan-etal-2021-bioelectra}
Kanakarajan, K.~r., Kundumani, B., and Sankarasubbu, M.
\newblock (2021).
\newblock {B}io{ELECTRA}:pretrained biomedical text encoder using
  discriminators.
\newblock In {\em Proceedings of the 20th Workshop on Biomedical Language
  Processing}, pages 143--154, Online, June. Association for Computational
  Linguistics.

\bibitem[\protect\citename{Kemper \bgroup et al.\egroup
  }2010]{kemper2010pathtext}
Kemper, B., Matsuzaki, T., Matsuoka, Y., Tsuruoka, Y., Kitano, H., Ananiadou,
  S., and Tsujii, J.
\newblock (2010).
\newblock Pathtext: a text mining integrator for biological pathway
  visualizations.
\newblock {\em Bioinformatics}, 26(12):i374--i381.

\bibitem[\protect\citename{Keymanesh \bgroup et al.\egroup
  }2021]{DBLP:journals/corr/abs-2109-14638}
Keymanesh, M., Elsner, M., and Parthasarathy, S.
\newblock (2021).
\newblock Privacy policy question answering assistant: {A} query-guided
  extractive summarization approach.
\newblock {\em CoRR}, abs/2109.14638.

\bibitem[\protect\citename{Krallinger \bgroup et al.\egroup
  }2017]{krallinger2017biocreative}
Krallinger, M., P{\'e}rez-P{\'e}rez, M., P{\'e}rez-Rodr{\'\i}guez, G.,
  Blanco-M{\'\i}guez, A., Fdez-Riverola, F., Capella-Gutierrez, S.,
  Louren{\c{c}}o, A., and Valencia, A.
\newblock (2017).
\newblock The biocreative v. 5 evaluation workshop: tasks, organization,
  sessions and topics.

\bibitem[\protect\citename{Landis and Koch}1977]{landis1977measurement}
Landis, J.~R. and Koch, G.~G.
\newblock (1977).
\newblock The measurement of observer agreement for categorical data.
\newblock {\em biometrics}, pages 159--174.

\bibitem[\protect\citename{Lauriola \bgroup et al.\egroup
  }2021]{lauriola2021learning}
Lauriola, I., Aiolli, F., Lavelli, A., and Rinaldi, F.
\newblock (2021).
\newblock Learning adaptive representations for entity recognition in the
  biomedical domain.
\newblock {\em Journal of biomedical semantics}, 12(1):1--13.

\bibitem[\protect\citename{Lee \bgroup et al.\egroup }2020]{lee2020biobert}
Lee, J., Yoon, W., Kim, S., Kim, D., Kim, S., So, C.~H., and Kang, J.
\newblock (2020).
\newblock Biobert: a pre-trained biomedical language representation model for
  biomedical text mining.
\newblock {\em Bioinformatics}, 36(4):1234--1240.

\bibitem[\protect\citename{Lev \bgroup et al.\egroup }2019]{lev2019talksumm}
Lev, G., Shmueli-Scheuer, M., Herzig, J., Jerbi, A., and Konopnicki, D.
\newblock (2019).
\newblock Talksumm: A dataset and scalable annotation method for scientific
  paper summarization based on conference talks.
\newblock {\em arXiv preprint arXiv:1906.01351}.

\bibitem[\protect\citename{Lewis \bgroup et al.\egroup }2019]{lewis2019bart}
Lewis, M., Liu, Y., Goyal, N., Ghazvininejad, M., Mohamed, A., Levy, O.,
  Stoyanov, V., and Zettlemoyer, L.
\newblock (2019).
\newblock Bart: Denoising sequence-to-sequence pre-training for natural
  language generation, translation, and comprehension.
\newblock {\em arXiv preprint arXiv:1910.13461}.

\bibitem[\protect\citename{Lin}2004]{lin2004rouge}
Lin, C.-Y.
\newblock (2004).
\newblock Rouge: A package for automatic evaluation of summaries.
\newblock In {\em Text summarization branches out}, pages 74--81.

\bibitem[\protect\citename{Liu and Lapata}2019]{liu2019text}
Liu, Y. and Lapata, M.
\newblock (2019).
\newblock Text summarization with pretrained encoders.
\newblock {\em arXiv preprint arXiv:1908.08345}.

\bibitem[\protect\citename{Loshchilov and
  Hutter}2017]{DBLP:journals/corr/abs-1711-05101}
Loshchilov, I. and Hutter, F.
\newblock (2017).
\newblock Fixing weight decay regularization in adam.
\newblock {\em CoRR}, abs/1711.05101.

\bibitem[\protect\citename{Miwa \bgroup et al.\egroup }2013]{miwa2013method}
Miwa, M., Ohta, T., Rak, R., Rowley, A., Kell, D.~B., Pyysalo, S., and
  Ananiadou, S.
\newblock (2013).
\newblock A method for integrating and ranking the evidence for biochemical
  pathways by mining reactions from text.
\newblock {\em Bioinformatics}, 29(13):i44--i52.

\bibitem[\protect\citename{Mulyar \bgroup et al.\egroup }2021]{mulyar2021mt}
Mulyar, A., Uzuner, O., and McInnes, B.
\newblock (2021).
\newblock Mt-clinical bert: scaling clinical information extraction with
  multitask learning.
\newblock {\em Journal of the American Medical Informatics Association},
  28(10):2108--2115.

\bibitem[\protect\citename{Narayan \bgroup et al.\egroup }2018]{narayan2018don}
Narayan, S., Cohen, S.~B., and Lapata, M.
\newblock (2018).
\newblock Don't give me the details, just the summary! topic-aware
  convolutional neural networks for extreme summarization.
\newblock {\em arXiv preprint arXiv:1808.08745}.

\bibitem[\protect\citename{Papanikolaou and
  Pierleoni}2020]{papanikolaou2020dare}
Papanikolaou, Y. and Pierleoni, A.
\newblock (2020).
\newblock Dare: Data augmented relation extraction with gpt-2.
\newblock {\em arXiv preprint arXiv:2004.13845}.

\bibitem[\protect\citename{Peng \bgroup et al.\egroup }2019]{peng2019transfer}
Peng, Y., Yan, S., and Lu, Z.
\newblock (2019).
\newblock Transfer learning in biomedical natural language processing: an
  evaluation of bert and elmo on ten benchmarking datasets.
\newblock {\em arXiv preprint arXiv:1906.05474}.

\bibitem[\protect\citename{Phan \bgroup et al.\egroup }2021]{phan2021scifive}
Phan, L.~N., Anibal, J.~T., Tran, H., Chanana, S., Bahadroglu, E., Peltekian,
  A., and Altan-Bonnet, G.
\newblock (2021).
\newblock Scifive: a text-to-text transformer model for biomedical literature.
\newblock {\em arXiv preprint arXiv:2106.03598}.

\bibitem[\protect\citename{Poon \bgroup et al.\egroup }2014]{poon2014literome}
Poon, H., Quirk, C., DeZiel, C., and Heckerman, D.
\newblock (2014).
\newblock Literome: Pubmed-scale genomic knowledge base in the cloud.
\newblock {\em Bioinformatics}, 30(19):2840--2842.

\bibitem[\protect\citename{Pu \bgroup et al.\egroup }2021]{Pu2021LearningCM}
Pu, A., Chung, H.~W., Parikh, A.~P., Gehrmann, S., and Sellam, T.
\newblock (2021).
\newblock Learning compact metrics for mt.
\newblock In {\em EMNLP}.

\bibitem[\protect\citename{Radford \bgroup et al.\egroup
  }2019]{radford2019language}
Radford, A., Wu, J., Child, R., Luan, D., Amodei, D., and Sutskever, I.
\newblock (2019).
\newblock Language models are unsupervised multitask learners.

\bibitem[\protect\citename{Raffel \bgroup et al.\egroup }2020a]{2020t5}
Raffel, C., Shazeer, N., Roberts, A., Lee, K., Narang, S., Matena, M., Zhou,
  Y., Li, W., and Liu, P.~J.
\newblock (2020a).
\newblock Exploring the limits of transfer learning with a unified text-to-text
  transformer.
\newblock {\em Journal of Machine Learning Research}, 21(140):1--67.

\bibitem[\protect\citename{Raffel \bgroup et al.\egroup
  }2020b]{raffel2020exploring}
Raffel, C., Shazeer, N., Roberts, A., Lee, K., Narang, S., Matena, M., Zhou,
  Y., Li, W., and Liu, P.~J.
\newblock (2020b).
\newblock Exploring the limits of transfer learning with a unified text-to-text
  transformer.
\newblock {\em Journal of Machine Learning Research}, 21:1--67.

\bibitem[\protect\citename{Sellam \bgroup et al.\egroup
  }2020]{sellam2020bleurt}
Sellam, T., Das, D., and Parikh, A.~P.
\newblock (2020).
\newblock Bleurt: Learning robust metrics for text generation.
\newblock {\em arXiv preprint arXiv:2004.04696}.

\bibitem[\protect\citename{Shahbazi \bgroup et al.\egroup
  }2020]{shahbazi2020relation}
Shahbazi, H., Fern, X., Ghaeini, R., and Tadepalli, P.
\newblock (2020).
\newblock Relation extraction with explanation.
\newblock In {\em Proceedings of the 58th Annual Meeting of the Association for
  Computational Linguistics}, pages 6488--6494.

\bibitem[\protect\citename{Strötgen and Gertz}2012]{Martin-90}
Strötgen, J. and Gertz, M.
\newblock (2012).
\newblock Temporal tagging on different domains: Challenges, strategies, and
  gold standards.
\newblock In Nicoletta Calzolari~(Conference Chair), et~al., editors, {\em
  Proceedings of the Eight International Conference on Language Resources and
  Evaluation (LREC'12)}, pages 3746--3753, Istanbul, Turkey, may. European
  Language Resource Association (ELRA).

\bibitem[\protect\citename{Subramani \bgroup et al.\egroup
  }2015]{subramani2015hpiminer}
Subramani, S., Kalpana, R., Monickaraj, P.~M., and Natarajan, J.
\newblock (2015).
\newblock Hpiminer: A text mining system for building and visualizing human
  protein interaction networks and pathways.
\newblock {\em Journal of Biomedical Informatics}, 54:121--131.

\bibitem[\protect\citename{Thayaparan \bgroup et al.\egroup
  }2020]{thayaparan2020survey}
Thayaparan, M., Valentino, M., and Freitas, A.
\newblock (2020).
\newblock A survey on explainability in machine reading comprehension.
\newblock {\em arXiv preprint arXiv:2010.00389}.

\bibitem[\protect\citename{Valenzuela-Esc{\'a}rcega \bgroup et al.\egroup
  }2018]{Escarcega:2018}
Valenzuela-Esc{\'a}rcega, M.~A., Babur, {\"O}., Hahn-Powell, G., Bell, D.,
  Hicks, T., Noriega-Atala, E., Wang, X., Surdeanu, M., Demir, E., and
  Morrison, C.~T.
\newblock (2018).
\newblock Large-scale automated machine reading discovers new cancer driving
  mechanisms.
\newblock {\em Database: The Journal of Biological Databases and Curation}.

\bibitem[\protect\citename{Wang \bgroup et al.\egroup
  }2019]{wang2019paperrobot}
Wang, Q., Huang, L., Jiang, Z., Knight, K., Ji, H., Bansal, M., and Luan, Y.
\newblock (2019).
\newblock Paperrobot: Incremental draft generation of scientific ideas.
\newblock {\em arXiv preprint arXiv:1905.07870}.

\bibitem[\protect\citename{Yao \bgroup et al.\egroup }2010]{yao2010collective}
Yao, L., Riedel, S., and McCallum, A.
\newblock (2010).
\newblock Collective cross-document relation extraction without labelled data.
\newblock In {\em Proceedings of the 2010 Conference on Empirical Methods in
  Natural Language Processing}, pages 1013--1023.

\bibitem[\protect\citename{Yasunaga \bgroup et al.\egroup
  }2019]{yasunaga2019scisummnet}
Yasunaga, M., Kasai, J., Zhang, R., Fabbri, A.~R., Li, I., Friedman, D., and
  Radev, D.~R.
\newblock (2019).
\newblock Scisummnet: A large annotated corpus and content-impact models for
  scientific paper summarization with citation networks.
\newblock In {\em Proceedings of the AAAI Conference on Artificial
  Intelligence}, volume~33, pages 7386--7393.

\bibitem[\protect\citename{Yuan \bgroup et al.\egroup
  }2020]{yuan2020constructing}
Yuan, J., Jin, Z., Guo, H., Jin, H., Zhang, X., Smith, T., and Luo, J.
\newblock (2020).
\newblock Constructing biomedical domain-specific knowledge graph with minimum
  supervision.
\newblock {\em Knowledge and Information Systems}, 62(1):317--336.

\bibitem[\protect\citename{Zhao \bgroup et al.\egroup }2020]{zhao2020ontosem}
Zhao, L., Wang, J., Cheng, L., and Wang, C.
\newblock (2020).
\newblock Ontosem: an ontology semantic representation methodology for
  biomedical domain.
\newblock In {\em 2020 IEEE International Conference on Bioinformatics and
  Biomedicine (BIBM)}, pages 523--527. IEEE.

\end{thebibliography}

\end{document}